\documentclass[11pt]{article}

\usepackage[preprint]{acl}

\usepackage{times}
\usepackage{latexsym}

\usepackage[T1]{fontenc}

\usepackage[utf8]{inputenc}

\usepackage{microtype}

\usepackage{inconsolata}

\usepackage{graphicx}
\usepackage{bbm}
\usepackage{amsmath}    
\usepackage{amssymb}      
\usepackage{booktabs}        

\usepackage{algorithm}     
\usepackage{algpseudocode}  
\algrenewcommand\algorithmicrequire{\textbf{Require:}}
\algrenewcommand\algorithmicensure{\textbf{Ensure:}}
\algrenewcommand\alglinenumber[1]{\scriptsize #1:}

\usepackage{float}      
\usepackage{placeins}

\usepackage{multirow}
\usepackage[table]{xcolor}

\definecolor{bestbg}{RGB}{255,200,200}    
\definecolor{secondbg}{RGB}{200,220,255}  

\usepackage{pifont}

\usepackage[bottom]{footmisc}

\title{RADS: Reinforcement Learning-Based Sample Selection Improves Transfer Learning in Low-resource and Imbalanced Clinical Settings}

\author{
  \textbf{Wei Han\textsuperscript{1}},
  \textbf{David Martinez\textsuperscript{1}},
  \textbf{Anna Khanina\textsuperscript{3,4,5}},
  \textbf{Lawrence Cavedon\textsuperscript{1}},
  \textbf{Karin Verspoor\textsuperscript{1,2,3}\thanks{Corresponding author: \href{mailto:karin.verspoor@rmit.edu.au}{karin.verspoor@rmit.edu.au}}}
\\
\\
  \textsuperscript{1}School of Computing Technologies, RMIT University\\
  \textsuperscript{2}School of Computing and Information Systems, The University of Melbourne\\
  \textsuperscript{3}National Centre for Infections in Cancer, Melbourne\\
  \textsuperscript{4}Department of Infectious Disease, Peter MacCallum Cancer Centre\\
  \textsuperscript{5}Sir Peter MacCallum Department of Oncology, The University of Melbourne
  }

\begin{document}
\maketitle
\begin{abstract}
A common strategy in transfer learning is few shot fine-tuning, but its success is highly dependent on the quality of samples selected as training examples. Active learning methods such as uncertainty sampling and diversity sampling can select useful samples. However, under extremely low-resource and class-imbalanced conditions, they often favor outliers rather than truly informative samples, resulting in degraded performance. In this paper, we introduce \textbf{RADS} (\textbf{R}einforcement \textbf{A}daptive \textbf{D}omain \textbf{S}ampling), a robust sample selection strategy using reinforcement learning (RL) to identify the most informative samples. Experimental evaluations on several real world clinical datasets show our sample selection strategy enhances model transferability while maintaining robust performance under extreme class imbalance compared to traditional methods.
Our code is open-sourced on GitHub\footnote{\url{https://github.com/Wei-0808/RADS}}.
\end{abstract}

\section{Introduction}

Maximizing the utility of limited data is a crucial focus of Natural Language Processing (NLP) research in domains such as clinical texts where acquiring large amounts of gold standard data may be difficult due to data restrictions and the relative rarity of many disease conditions. The high cost of annotation in such highly specialized domains further limits availability of labeled data. Yet, the effectiveness of NLP techniques in healthcare heavily relies on the quality of annotated datasets, particularly because clinical data contains specialized symbols, abbreviations, and medical jargon \cite{touvron2023llama, liu2024deepseek}. 

Transfer Learning (TL) \cite{tan2018survey}, in which knowledge learned from a task is reused to boost performance on a different but related (target) task, has shown effectiveness across various machine learning applications \cite{weiss2016survey} and opens new avenues for addressing low-resource scenarios. 
Previous works have attempted to leverage pretrained embeddings \cite{maimaiti2021enriching} and few shot examples \cite{alyafeai2020survey} to facilitate transfer learning in NLP. However, when the target task offers very few labeled instances, these approaches may generate unreliable outputs. This is an especially acute problem in healthcare, where reliability is paramount.

\begin{figure}[t]
  \centering
  \includegraphics[width=0.955\columnwidth]{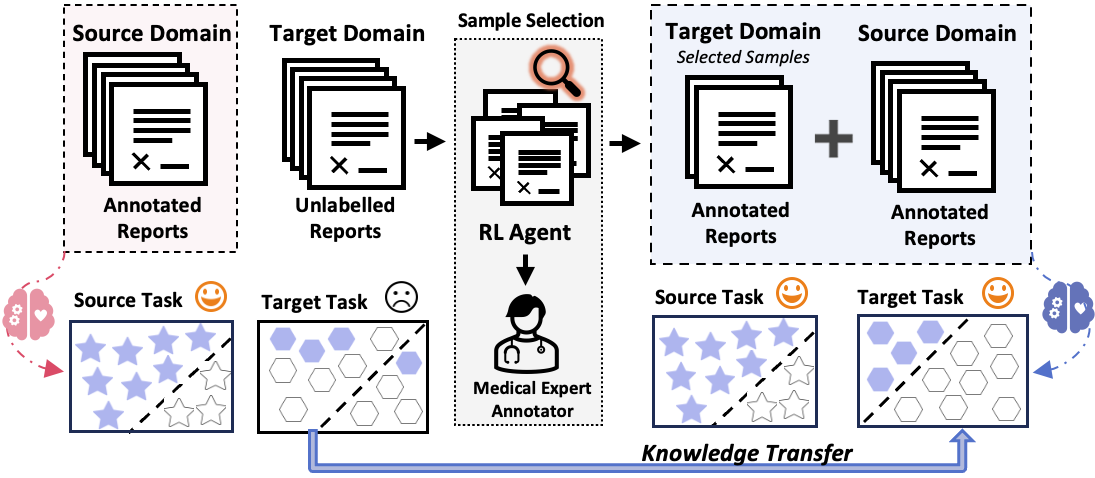}
  \caption{RL-based active sampling for transfer learning from source domain to target domain. Domain shift reduces zero-shot generalization from the source-trained model to the target domain. Our sample selection strategy uses RL to identify key samples from the target domain. By jointly fine-tuning on the selected target samples with the source data, the model achieves good performance on both domains.}
  \label{fig:RL}
\end{figure}

Class imbalance \cite{johnson2019survey} is another challenge for low-resource settings. In clinical datasets, there is often a scarcity of positive cases due to the low prevalence of many conditions, making such instances both highly valuable and limited in number. At the same time, differences in data collection protocols can lead some disease datasets to contain a very high proportion of positive samples. These extreme disparities in class distribution further hinder the transferability of NLP models across clinical datasets.


Clinical documentation is heterogeneous, reflecting diverse investigations including CT and PET scans, or cytology and histopathology analysis. Although disease detection cues appear across these document types, their content structures, terminology, and linguistic expressions can vary greatly. CT and PET scan reports primarily emphasize imaging-based findings \cite{townsend2004pet}, whereas cytology and histopathology reports focus on cellular and tissue-level observations \cite{jensen2021histopathology}.

Previous works have shown the efficacy of NLP techniques for disease detection from clinical reports. However, models fine-tuned on one report type show clear performance degradation when applied to another~\cite{han2025automated}. While disease-related signals overlap to some extent across different document types, existing disease detection models still fall short of human performance in transferring knowledge between them. As the preparation of gold standard annotated datasets for training is time-consuming, it is therefore important to explore effective knowledge transfer strategies from existing datasets to new but similar tasks. This not only improves annotation efficiency but also enhances the models' adaptability in dealing with variations in task settings.

In this work, we propose RADS (Reinforcement Adaptive Domain Sampling), a robust strategy for knowledge transfer between related but distinct sources. 
Following the active learning paradigm \cite{fu2013survey}, we enhance transfer learning by identifying and selecting the most relevant samples for few-shot fine-tuning as shown in Figure~\ref{fig:RL}. First, we employ an RL-based agent to identify the most informative samples within the target dataset. 
These selected samples are then annotated by medical experts and incorporated into the fine-tuning process. By jointly fine-tuning the model on the source dataset and the newly annotated target samples, the model is able to preserve strong performance on the source domain while achieving improved generalization to the target domain.
We evaluated this approach across multiple real world clinical datasets. Experimental results show that our method improves both the adaptability and performance of disease detection between different sources. 
In the context of transfer learning, this technique offers a promising way to both reduce annotation effort and enhance model robustness in low-resource and class-imbalanced settings.

Our contributions are summarized as follows:%
\begin{itemize}
    \item This work addresses the challenges posed by low-resource and class-imbalance scenarios in disease detection across heterogeneous clinical report types from real-world clinical data sources.
    \item We propose RADS, a robust RL-based sample selection strategy tailored to scenarios with both data scarcity and class imbalance.
    \item Extensive experiments on several clinical datasets confirm that our transfer learning approach is more effective between similar but different sources, even under low-resource and class-imbalanced conditions.
\end{itemize}

\section{Related Work}
With high-quality annotated datasets, NLP methods have shown promising results in disease detection. Based on the concept features relevant to diseases, dictionary-based detection approaches and classical machine learning have shown effective performance \cite{rozova2023detecting, martinez2015automatic}. Bag-of-words models have also been utilized, often combined with machine learning techniques to further enhance accuracy and scalability in disease detection \cite{cury2021natural,lopez2020covid}. 
Recently, large language models (LLMs), such as BioBERT \cite{lee2020biobert} and ClinicalBERT \cite{huang2019clinicalbert}, pre-trained on large biomedical corpora, have improved contextual understanding in clinical texts \cite{consoli2024epidemic, han2025automated}.

Low resource settings remain challenging for NLP tasks. 
Few-shot fine-tuning \cite{NEURIPS2020_1457c0d6, gu-etal-2022-ppt, liu2022few}, where large pre-trained models are adapted using only a small number of labeled examples, has shown promising results. Selecting effective few-shot samples is critical, and active learning strategies such as uncertainty sampling \cite{nguyen2022measure} and diversity sampling \cite{yang2015multi} are often employed.
However, these methods typically optimize a single metric, and under domain shift, tend to select distributional outliers rather than truly informative samples \cite{gonsior2024comparing}. Reinforcement Learning (RL) \cite{fang-etal-2017-learning,liu2024review} offers a potential solution by optimizing more flexible and adaptive sample selection policies, thereby improving robustness in different contexts. 

Class imbalance is especially crucial in low-resource clinical NLP tasks \cite{ghosh2024class}. 
Data-level approaches, such as oversampling minority classes \cite{hairani2024addressing}
and undersampling majority classes \cite{yang2024impact}, are typically used to balance class distributions. Algorithm-level methods, such as cost-sensitive learning \cite{araf2024cost} and focal loss adjustments \cite{aljohani2023novel}, aim to direct model attention towards underrepresented classes, thereby improving model performance in class-imbalanced settings. 

\section{Methodology}
\label{sec:method}

\subsection{Problem Setup and Overview}
We study low-resource and class-imbalanced transfer learning between heterogeneous clinical report datasets: a fully labeled source dataset $\mathcal{D}_s$ and an unlabeled target dataset $\mathcal{U}_t$. Although the two datasets (domains) share some similar clinical knowledge, distribution shift and differences in label distribution make direct transfer challenging. 

We formulate cross-domain adaptation as a budgeted active learning problem: given an annotation budget $B \ll N_t$, where $N_t$ is the target pool size, our goal is to select a small but high-utility subset $\mathcal{Q}\subset\mathcal{U}_t$. The selected samples are then annotated and merged with $\mathcal{D}_s$ to form an expanded training set. With supervised fine-tuning on the final dataset, the knowledge can be effectively transferred and the model performance across both domains also improved.

\begin{figure}[t]
  \centering
  \includegraphics[width=0.85\linewidth]{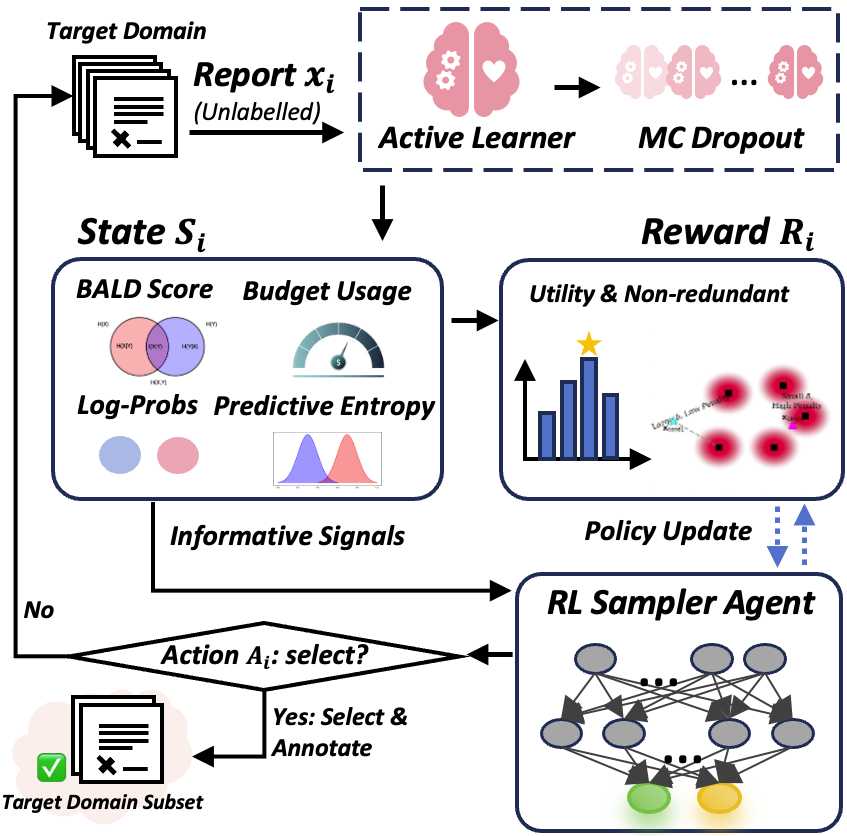}  
  \caption{RADS framework for RL-based active sampling under domain shift. The active learner is fine-tuned on the source domain, and MC dropout is used to score unlabeled target reports and construct informativeness signals (state). An RL sampler then selects a subset for annotation by maximizing the reward, producing a target set for joint fine-tuning with the source data.}

  \label{fig:method}
\end{figure}

The overall framework of RADS is shown in Figure \ref{fig:method}. Our approach consists of three stages: (1) we train an active learner on $\mathcal{D}_s$ and compute informativeness signals for $\mathcal{U}_t$ via Monte-Carlo (MC) dropout; (2) we define a prior-aware utility that combines BALD-based mutual information ~\cite{houlsby2011bayesian} with pseudo-label class weighting to explicitly control the quality of selected samples for transfer learning under severe class imbalance; and (3) we train a reinforcement learning sampler to select samples that maximize prior-aware utility while discouraging redundant selections. The pseudocode for this part is provided in Appendix \ref{sec:algo}.

\subsection{Active Learner}

We first fine-tune a lightweight classifier $f_{\phi}$ on the labeled source dataset $\mathcal{D}_s$. For each unlabeled target report in training pool $x\in\mathcal{U}_t$, we estimate epistemic uncertainty via MC dropout \cite{gal2016dropout}. Specifically, we keep dropout activated at inference time and perform $K$ stochastic forward passes. Each pass corresponds to sampling a dropout mask, yielding a sampled set of network weights $\mathbf{w}_k$ and a predictive distribution:
\begin{equation}
  p_k(y \mid x) \;=\; \mathrm{softmax}\!\left(f_{\phi}(x; \mathbf{w}_k)\right)
\end{equation}
Aggregating these $K$ stochastic predictions approximates the posterior predictive distribution. We compute the MC predictive mean as:
\begin{equation}
  \bar{p}(y \mid x) \;=\; \frac{1}{K}\sum_{k=1}^{K} p_k(y \mid x)
\end{equation}

In addition, we retain the mean log-probability vector $\bar{\ell}(x)=\log \bar{p}(\cdot\mid x)$, which serves as a representation for redundancy estimation in our RL-based sampler (Section~\ref{sec:rl}).

Based on this active learner, we also define a pseudo label $\hat{y}(x)=\arg\max_y \bar{p}(y\mid x)$ and estimate the predicted target class prior:
\begin{equation}
\begin{aligned}
  \hat{\pi}_{+} \;&=\; \frac{1}{N_t}\sum_{x\in\mathcal{U}_t} \mathbbm{1}\!\left[\hat{y}(x)=1\right], \\
  \hat{\pi}_{-} \;&=\; 1-\hat{\pi}_{+}.
\end{aligned}
\end{equation}
These priors let us correct selection bias when the pool is imbalanced or the source-trained model is miscalibrated on the target domain.

\subsection{BALD Signal}
To score informativeness for unlabeled target-domain samples, we use BALD, which quantifies the mutual information between the predicted label and the model parameters. 
Let $H(\cdot)$ denote entropy. For each $x\in\mathcal{U}_t$, we compute:
\begin{align}
  \mathrm{PE}(x) &= H\!\left(\bar{p}(\cdot \mid x)\right), \\
  \mathrm{EE}(x) &= \frac{1}{K}\sum_{k=1}^{K} H\!\left(p_k(\cdot \mid x)\right), \\
  \mathrm{MI}(x) &= \mathrm{PE}(x) - \mathrm{EE}(x).
\end{align}

Here, $\mathrm{MI}(x)$ is the BALD score. It is large when the predictive distribution is uncertain overall (high $\mathrm{PE}$) while individual stochastic models are relatively confident but disagree with each other (low $\mathrm{EE}$). We normalize $\mathrm{MI}(x)$ to $[0,1]$ over $\mathcal{U}_t$, denoted as $\widetilde{\mathrm{MI}}(x)$.

We treat samples with high $\widetilde{\mathrm{MI}}(x)$ as informative and assign them higher utility in our selection policy. Prioritizing these samples for annotation is expected to reduce the model's uncertainty and improve transfer to the target domain. 

\begin{table*}[h]
\centering
\small
\setlength{\tabcolsep}{5pt}
\renewcommand{\arraystretch}{1}
\begin{tabular}{lccc}
\toprule
\textbf{Attribute} & \textbf{CHIFIR} & \textbf{PIFIR} & \textbf{MIMIC-CXR (subset)} \\
\midrule
Report Type & Cytology / Histopathology & PET--CT Radiology & Chest X-ray Radiology \\
Target Disease & Invasive Fungal Infection (IFI) & Invasive Fungal Infection (IFI) & Pneumonia \\
Label Type & Gold (manual) & Gold (manual) & Silver (auto-derived) \\
Class distribution (P/N) & 14\% / 86\% & 69\% / 31\% & 39\% / 61\% \\
Dataset Size & 283 reports (small) & 201 reports (small) & 493 reports (medium)\\
\bottomrule
\end{tabular}
\caption{Key attributes of the three datasets used in this study. P/N = positive/negative class proportions}

\label{tab:dataset-summary}
\end{table*}

\subsection{Prior-Aware Utility for Sample Selection}
\label{sec:utility}
Selecting the top-$B$ uncertain samples can sometimes produce an extreme class skew. This often happens under domain shift and severe class imbalance, where the source-trained active learner may predict biased pseudo labels on the target domain. To control the selected class mixture, we introduce a prior-aware utility. 
We define class weights using the estimated prior:
\begin{equation}
\begin{aligned}
  w_{+} \;&=\; \frac{\rho}{\mathrm{clip}(\hat{\pi}_{+})}, \\
  w_{-} \;&=\; \frac{1-\rho}{1-\mathrm{clip}(\hat{\pi}_{+})}.
\end{aligned}
\end{equation}
where $\mathrm{clip}(\cdot)$ clamps probabilities away from $\{0,1\}$ for stability and $\rho$ is a hyperparameter that trades off class-balance control and informativeness. We then define the utility:
\begin{equation}
  u(x) \;=\; \widetilde{\mathrm{MI}}(x)\cdot
  \begin{cases}
    w_{+}, & \hat{y}(x)=1,\\
    w_{-}, & \hat{y}(x)=0.
  \end{cases}
  \label{eq:utility}
\end{equation}
This utility favors informative samples and shifts selection toward the desired class ratio.

\subsection{RL-based Sample Selection Strategy}
\label{sec:rl}
At each step $t$, the sampler agent observes the current candidate $x_t$ and decides whether to select or discard. An episode ends when $B$ samples are selected or the pool is exhausted.

\paragraph{State.}
For each candidate $x_t$, the state vector combines the active learner signals and a budget progress term:
\begin{equation}
  s_t \;=\; \Big[\;\bar{\ell}(x_t);\;\mathrm{PE}(x_t);\;\mathrm{MI}(x_t);\;|S_t|/B\;\Big]
\end{equation}
where $\bar{\ell}(x_t)$ is the mean log-probability vector computed from MC dropout; $\mathrm{PE}(x_t)$ is the predictive entropy; $\mathrm{MI}(x_t)$ is the BALD score; and $|S_t|/B$ indicates the fraction of the annotation budget already consumed, with $S_t$ denoting the set of selected samples so far. In our binary setting, $\bar{\ell}(x_t)\in\mathbb{R}^{2}$, hence the overall dimension of the state vector is 5.

\paragraph{Reward.}
Our reward encourages the agent to select samples that are both (i) informative for learning under class imbalance and (ii) non-redundant with respect to previously selected instances. Specifically, when the agent selects the current candidate ($a_t=1$) and the budget is not yet exhausted ($|S_t|<B$), we define:
\begin{equation}
  r_t \;=\; u(x_t) \;-\; \lambda \cdot \mathrm{Red}(x_t, S_t)
  \label{eq:reward}
\end{equation}
and set $r_t=0$ otherwise. Here, $u(x_t)$ is the prior-aware utility (Section~\ref{sec:utility}) and $\lambda$ controls the strength of the diversity regularization.

To discourage selecting near-duplicate samples, we measure redundancy in the active learner's predictive representation space. For a candidate $x$ and the current selected set $S$, we first compute the distance to its nearest selected neighbor:
\begin{equation}
\delta(x,S)
=
\begin{cases}
+\infty, & |S|=0,\\[2pt]
\min\limits_{x'\in S}\left\lVert \bar{\ell}(x)-\bar{\ell}(x')\right\rVert_2, & \text{otherwise.}
\end{cases}
\end{equation}
We then convert this distance into a bounded redundancy score:
\begin{equation}
\mathrm{Red}(x,S)
=
\begin{cases}
0, & |S|=0,\\[2pt]
\dfrac{1}{1+\delta(x,S)}, & \text{otherwise.}
\end{cases}
\label{eq:redundancy}
\end{equation}
This definition yields a larger penalty when $x$ is very close to an existing selection (small $\delta$), and a smaller penalty when $x$ is far away (large $\delta$). As a result, the agent is encouraged to select diverse samples while still prioritizing high-utility ones.

\paragraph{Dueling DQN Sampler Agent.}
We learn a Q-function $Q_{\theta}(s,a)$ with a {dueling DQN architecture \cite{wang2016dueling} and optimize it via the standard DQN objective \cite{mnih2015human}. We maintain an experience replay buffer $\mathcal{B}$ and a target network $Q_{\theta^-}$. At each gradient step, we minimize the temporal-difference loss:
\begin{equation}
\begin{aligned}
\mathcal{L}(\theta)
&=\mathbb{E}_{(s,a,r,s',d)\sim \mathcal{B}}
\Big[
\big(
Q_{\theta}(s,a) - y
\big)^2
\Big],\\
y
&= r+\gamma(1-d)\max_{a'}Q_{\theta^-}(s',a').
\end{aligned}
\label{eq:dqn_loss}
\end{equation}

where $\gamma$ is the discount factor and $d\in\{0,1\}$ indicates episode termination. We adopt $\epsilon$-greedy exploration with a decaying $\epsilon$ schedule, periodically synchronize $\theta^-$ with $\theta$, and finally use the learned policy $\pi(s)=\arg\max_a Q_{\theta}(s,a)$ to select $B$ samples from $\mathcal{U}_t$.

\section{Experimental Setup}

\subsection{Benchmark Datasets}

We chose three real world clinical datasets \cite{rozova2023chifir, rozova2025pifir, johnson2019mimic} as benchmarks in this study: the PET-CT Invasive Fungal Infection Reports corpus (PIFIR\footnote{Available for credentialed users at \url{https://physionet.org/content/pifir/1.0.0/ }}), the Cytology and Histopathology IFI Reports corpus (CHIFIR\footnote{Available for credentialed users at \url{https://physionet.org/content/corpus-fungal-infections/1.0.2/}}), and the MIMIC Chest X-ray corpus (MIMIC-CXR\footnote{Available for credentialed users at \url{https://physionet.org/content/mimic-cxr/2.1.0/}}). 

CHIFIR and PIFIR datasets are related to Invasive Fungal Infection (IFI), but the vocabulary used varies across them. The cytology and histopathology reports of the CHIFIR dataset assess tissue or fluid samples and describe the microscopic visualization of fungal organisms. The PET-CT reports from PIFIR assess metabolic activity and discuss the anatomical and morphological features of fungal lesions via PET imaging. 

To assess transfer beyond IFI and beyond pathology-style reports, we also include MIMIC-CXR, a corpus of chest X-ray reports. We construct a \textit{Pneumonia} subset by selecting the top 3,000 reports that are labeled as pneumonia by CheXpert's \cite{irvin2019chexpert} weak labels.
Although PIFIR and MIMIC-CXR both consist of radiology reports, they still differ greatly in reporting style and clinical phrasing. Moreover, pneumonia and IFI reflect distinct 
clinical contexts, further increasing the domain shift. Figure \ref{fig:wordclouds} shows differences in predominant clinical terms across the three datasets.

\begin{figure}[b]
  \centering
  \includegraphics[width=0.32\linewidth]{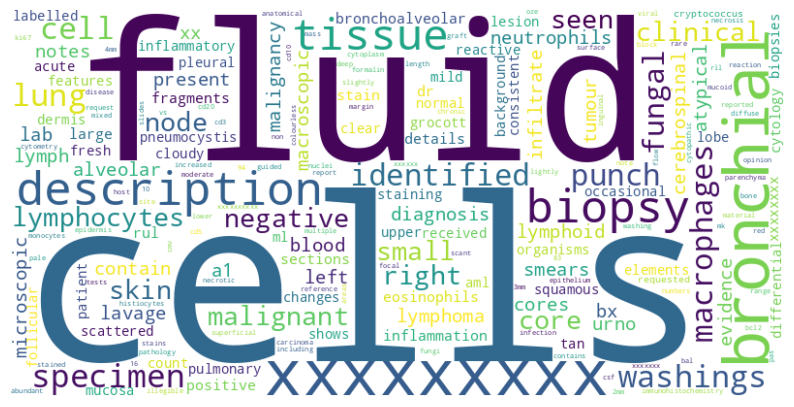} 
  \includegraphics[width=0.32\linewidth]{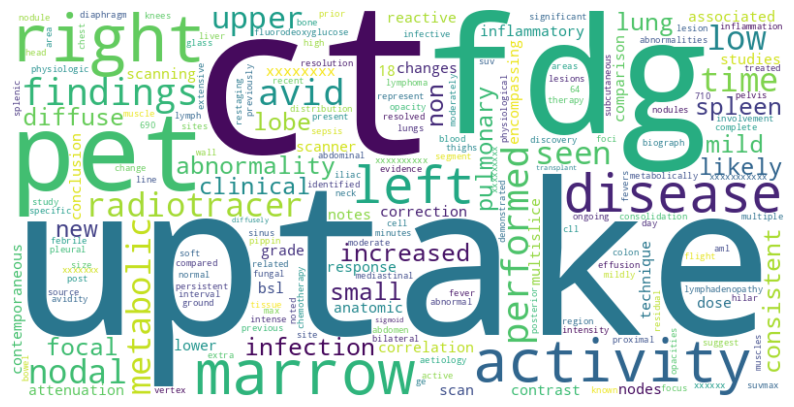}
  \includegraphics[width=0.32\linewidth]{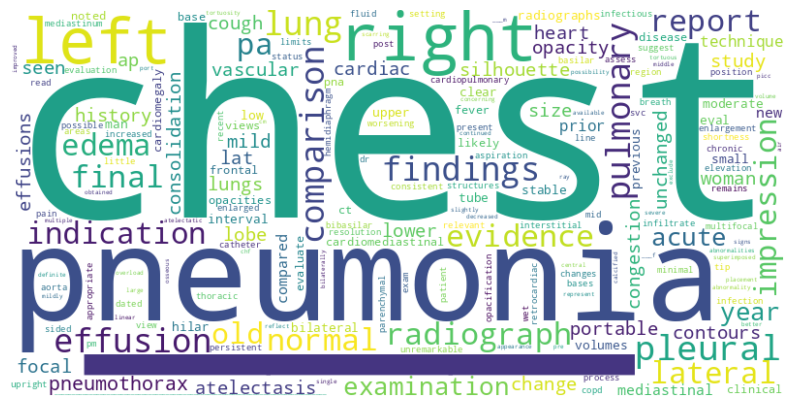}
  \caption{Word clouds for the CHIFIR (left), PIFIR (middle), and MIMIC-CXR (right) datasets. Word size corresponds to term frequency.}
  \label{fig:wordclouds}
\end{figure}

All three datasets exhibit class imbalance. CHIFIR and MIMIC-CXR are dominated by negative cases, whereas PIFIR is dominated by positive cases. Throughout the paper we focus on transferring from other sources to PIFIR, and we provide results of other directions in the Appendix.


Table \ref{tab:dataset-summary} summarizes the key characteristics of each dataset and highlights the challenges for transfer learning across them. Details of the dataset split are provided in Appendix~\ref{sec:repro}.


\subsection{Evaluation Metrics}

We evaluate performance using accuracy, F1 score, precision, recall, and ROC-AUC. Class imbalance in benchmark datasets makes the F1 score particularly important. Recall is also important, given that it is critical not to miss positive cases. 

\begin{table*}[t]
  \centering
  \small
  \renewcommand{\arraystretch}{1}       
  \setlength{\tabcolsep}{4.5pt}
  \setlength{\heavyrulewidth}{1.2pt}  
  \resizebox{\textwidth}{!}{
  \begin{tabular}{ll|cccc|cccc|cccc}
    \toprule
    \multicolumn{2}{c|}{\textbf{Transfer Learning to PIFIR}} 
      & \multicolumn{4}{c|}{\textbf{Performance on PIFIR}} 
      & \multicolumn{4}{c|}{\textbf{Performance on CHIFIR}} 
      & \multicolumn{4}{c}{\textbf{Performance on MIMIC-CXR}} \\
    \multicolumn{1}{c}{\textbf{Strategy}} 
      & \multicolumn{1}{c|}{\textbf{Datasets}}
      & Acc & F1 & P & R 
      & Acc & F1 & P & R 
      & Acc & F1 & P & R \\
    \midrule
    Baseline
      & PIFIR
      & 0.714 & 0.812 & 0.788 & 0.839
      & --   & --   & --   & --    
      & --   & --   & --   & -- \\ 
    \midrule
    \multirow{2}{*}{Zero-shot} 
      & CHIFIR
      & 0.357 & 0.229 & 1.000 & 0.129
      & 0.942 & 0.824 & 0.778 & 0.875 
      & --   & --   & --     & -- \\ 
      & MIMIC-CXR 
      & 0.738 & 0.841 & 0.763 & 0.935
      & --   & --   & --   & -- 
      & 0.859 & 0.811 & 0.833 & 0.789 \\
    \midrule
    \multirow{2}{*}{Full-shot}
      & CHIFIR + PIFIR
      & 0.857 & 0.900 & 0.931 & 0.871
      & 0.904 & 0.615 & 0.800 & 0.500
      & --  & --   & --   & -- \\
      & MIMIC-CXR + PIFIR
      & 0.833 & 0.889 & 0.875 & 0.903
      & --   & --   & --   & -- 
      & 0.848 & 0.800 & 0.811 & 0.789\\
    \bottomrule
  \end{tabular}
  }
  \caption{
  Performance comparison of ClinicalBERT under different transfer strategies. Zero-shot transfer refers to fine-tuning solely on the source dataset. We set this as our baseline. Full-shot transfer refers to jointly fine-tuning on both source and target datasets. We assume this represents the best possible transfer learning performance.
  }
  \label{tab:baseline}
\end{table*}

\subsection{Baselines}

We select the fine-tuned ClinicalBERT approach from previous work \cite{han2025automated} as the baseline. Table \ref{tab:baseline} shows the baseline results and reveals the challenges of knowledge transferability between these datasets. Models perform well when fine-tuned and evaluated on the same dataset. Without transfer learning, evaluation on a similar but still different dataset results in a clear performance drop. Although training on all datasets together can improve performance, it requires annotating all reports, which is labor-intensive.  Full reproducibility details can be found in Appendix \ref{sec:repro}.

\section{Experimental Results}

\subsection{Transfer Learning Performance}

We compare our method with several other active learning approaches to analyze the impact of different sample selection methods on knowledge transfer performance.
\noindent{}1) \textbf{Random Selection}: Randomly select samples from the unlabeled target domain. Each experiment is run five times to reduce variance and obtain more reliable results. We report the mean evaluation metrics over these five runs.
\noindent{}2) \textbf{Uncertainty-based Selection} \cite{nguyen2022measure}: Selects \(k \) samples by predictive uncertainty (lowest confidence) from the active learner.
\noindent{}3) \textbf{Diversity-based Selection} \cite{yuan-etal-2020-cold}: 
Selects the \(k \) most diverse samples by calculating the cosine distance between each report embedding in the unlabeled dataset and the embeddings in the labeled dataset. 
\noindent{}4) \textbf{LM-DPP Selection} \cite{wang-etal-2024-effective}: This method jointly models uncertainty and diversity using a Determinantal Point Process (DPP) kernel. Following the original work, we set the trade-off coefficient between uncertainty and diversity to 0.5 and select the subset of size \(k\) that maximizes the DPP objective for annotation.
\noindent{}5) \textbf{TAGCOS Selection} \cite{zhang2025tagcos}: A task-agnostic selection baseline that selects \(k\) samples according to its gradient-based selection criterion. 
 \noindent{}6) \textbf{BatchBALD Selection} \cite{kirsch2019batchbald}: Selects \(k \) samples using a batch acquisition strategy that extends BALD by maximizing joint mutual information under MC dropout. 
\begin{table*}[h]
  \centering
  \scriptsize
  \renewcommand{\arraystretch}{1.15}
  \setlength{\tabcolsep}{4pt}
  \resizebox{0.95\textwidth}{!}{%
    \begin{tabular}{c|ccccc|ccccc|cc}
      \toprule
      \multicolumn{13}{c}{\textbf{Knowledge Transfer from CHIFIR to PIFIR}} \\
      \midrule
      \multirow{2}{*}{\textbf{Strategy}}
        & \multicolumn{5}{c|}{\textbf{Performance on PIFIR}}
        & \multicolumn{5}{c|}{\textbf{Performance on CHIFIR}}
        & \multicolumn{2}{c}{\textbf{Transfer Gap}} \\
      & Accuracy & F1-score & Precision & Recall & ROC-AUC
        & Accuracy & F1-score & Precision & Recall & ROC-AUC
        & $\Delta$F1 & 95\% CI \\
      \midrule

      Random
        & 0.595 & 0.639 & 0.885 & 0.561 & 0.813
        & 0.927 & 0.746 & 0.805 & 0.700 & 0.938
        & -- & -- \\
      Uncertainty
        & 0.524 & 0.545 & 0.923 & 0.387 & 0.830
        & 0.942 & 0.824 & 0.778 & 0.875 & 0.977
        & 0.278 & [-0.037, 0.530] \\
      Diversity
        & 0.595 & 0.638 & 0.938 & 0.484 & 0.809
        & 0.942 & 0.800 & 0.857 & 0.750 & 0.974
        & 0.162 & [-0.167, 0.412] \\
      LM-DPP
        & 0.571 & 0.609 & 0.933 & 0.452 & 0.839
        & 0.904 & 0.615 & 0.800 & 0.500 & 0.977
        & 0.007 & [-0.418, 0.319] \\
      TAGCOS
        & 0.762 & 0.844 & 0.818 & 0.871 & 0.730
        & 0.942 & 0.824 & 0.778 & 0.875 & 0.972
        & -0.020 & [-0.310, 0.168] \\
      BatchBALD
        & 0.738 & 0.849 & 0.738 & 1.000 & 0.783
        & 0.885 & 0.500 & 0.750 & 0.375 & 0.946
        & -0.349 & [-0.833, -0.019] \\
      \rowcolor{gray!16}
      \textbf{RADS}
        & \textbf{0.810} & \textbf{0.871} & \textbf{0.871} & \textbf{0.871} & \textbf{0.833}
        & \textbf{0.923} & \textbf{0.750} & \textbf{0.750} & \textbf{0.750} & \textbf{0.977}
        & \textbf{-0.121} & \textbf{[-0.430, 0.100]} \\
      \bottomrule
    \end{tabular}
  }
  \caption{Transfer learning performance from CHIFIR to PIFIR with 5 samples selected in PIFIR under different sample selection strategies. $\Delta$F1 = F1(CHIFIR) $-$ F1(PIFIR). CI = Confidence Interval.}
  \label{tab:cp_transfer}
\end{table*}

\begin{table*}[h]
  \centering
  \scriptsize
  \renewcommand{\arraystretch}{1.15}
  \setlength{\tabcolsep}{4pt}
  \resizebox{0.95\textwidth}{!}{%
    \begin{tabular}{c|ccccc|ccccc|cc}
      \toprule
      \multicolumn{13}{c}{\textbf{Knowledge Transfer from MIMIC-CXR to PIFIR}} \\
      \midrule
      \multirow{2}{*}{\textbf{Strategy}}
        & \multicolumn{5}{c|}{\textbf{Performance on PIFIR}}
        & \multicolumn{5}{c|}{\textbf{Performance on MIMIC-CXR}}
        & \multicolumn{2}{c}{\textbf{Transfer Gap}} \\
      & Accuracy & F1-score & Precision & Recall & ROC-AUC
        & Accuracy & F1-score & Precision & Recall & ROC-AUC
        & $\Delta$F1 & 95\% CI \\
      \midrule

      Random
        & 0.766 & 0.848 & 0.808 & 0.842 & 0.806
        & 0.862 & 0.824 & 0.808 & 0.842 & 0.932
        & -- & -- \\
      Uncertainty
        & 0.738 & 0.831 & 0.794 & 0.871 & 0.636
        & 0.848 & 0.810 & 0.780 & 0.842 & 0.916
        & -0.021 & [-0.165, 0.126] \\
      Diversity
        & 0.738 & 0.845 & 0.750 & 0.968 & 0.616
        & 0.859 & 0.816 & 0.816 & 0.816 & 0.933
        & -0.029 & [-0.157, 0.102] \\
      LM-DPP
        & 0.738 & 0.831 & 0.794 & 0.871 & 0.636
        & 0.848 & 0.810 & 0.780 & 0.842 & 0.916
        & -0.021 & [-0.165, 0.126] \\
      TAGCOS
        & 0.738 & 0.836 & 0.778 & 0.903 & 0.795
        & 0.862 & 0.831 & 0.821 & 0.842 & 0.930
        & -0.005 & [-0.131, 0.139] \\
      BatchBALD
        & 0.762 & 0.848 & 0.800 & 0.903 & 0.827
        & 0.889 & 0.861 & 0.829 & 0.895 & 0.949
        & 0.012 & [-0.107, 0.141] \\
      \rowcolor{gray!16}
      \textbf{RADS}
        & \textbf{0.810} & \textbf{0.882} & \textbf{0.811} & \textbf{0.968} & \textbf{0.880}
        & \textbf{0.869} & \textbf{0.840} & \textbf{0.791} & \textbf{0.895} & \textbf{0.921}
        & \textbf{-0.043} & \textbf{[-0.153, 0.072]} \\
      \bottomrule
    \end{tabular}
  }
  \caption{Transfer learning performance from MIMIC-CXR to PIFIR with 2 samples selected in PIFIR under different sample selection strategies. $\Delta$F1 = F1(MIMIC-CXR) $-$ F1(PIFIR). CI = Confidence Interval.}
  \label{tab:mp_transfer}
\end{table*}





Table~\ref{tab:cp_transfer} reports transfer learning results from CHIFIR to PIFIR. Uncertainty-, diversity-, and LM-DPP-based selection yield comparable or worse performance than random sampling. While TAGCOS and BatchBALD attain relatively high F1 scores on PIFIR, their ROC-AUC is noticeably lower. In contrast, our method RADS achieves the best performance on PIFIR while maintaining competitive performance on the source domain (CHIFIR). This indicates strong sample efficiency, requiring only \(5/135 \approx 3.7\%\) of the target training set to obtain substantial transfer gains.

We further compare RADS with a prompt-guided LLM selection baseline \cite{jeong2025llmselect}, which uses an open-source medical LLM to score report and then selects the top-k reports for annotation. We report the CHIFIR to PIFIR transfer results in Appendix~\ref{app:llm_selection}. Although this baseline can retrieve some useful reports as the budget increases, it is highly unstable in the ultra-low-budget regime and remains less reliable than RADS overall. 

Table~\ref{tab:mp_transfer} reports transfer learning results from MIMIC-CXR to PIFIR. 
Other baselines provide limited gains and are often on par with or below random selection. RADS achieves better target performance (F1 on PIFIR = 0.882) and remains highly sample-efficient, requiring annotation of only \(2/135 \approx 1.5\%\) of the target training set.


We also conduct transfer learning experiments from PIFIR to CHIFIR. The results show that our method still achieves the best performance with only 8 samples selected from target dataset CHIFIR. More detailed discussion is shown in Appendix~\ref{sec:p->c}.


RADS consistently outperforms strong baselines, demonstrating superior sample-efficient transfer performance.
More robustness analysis under imbalanced settings appears in  Appendix \ref{sec:robust}.

\subsection{Learning Curves under Varying Budgets}

\begin{figure}[h]
  \centering
  \includegraphics[width=0.49\linewidth]{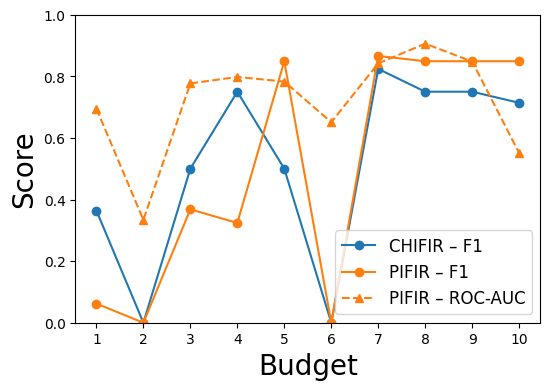} 
  \includegraphics[width=0.49\linewidth]{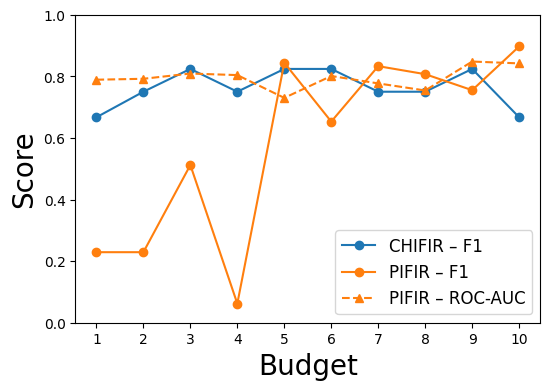}
  \caption{Transfer from CHIFIR to PIFIR 
  under baselines BatchBALD (left) and TAGCOS (right).}
  \label{fig:sample}
\end{figure}
\begin{figure}[h]
  \centering
  \includegraphics[width=0.48\linewidth]{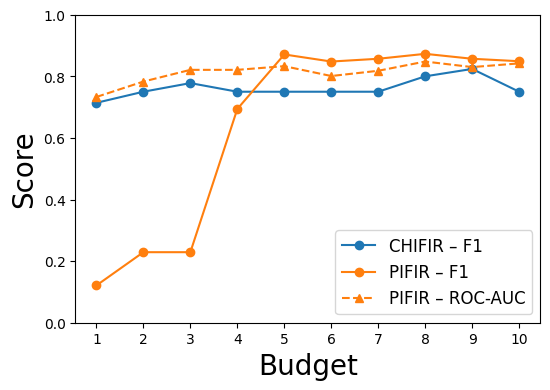}
  \includegraphics[width=0.49\linewidth]{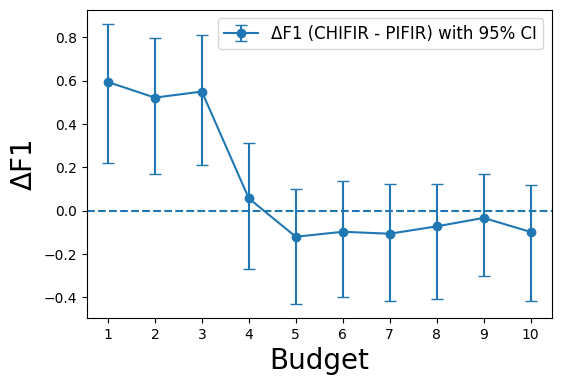}
  \caption{Transfer from CHIFIR to PIFIR 
  under our method RADS.}
  \label{fig:F1}
\end{figure}

We analyze the effect of annotation budget on transfer performance. 
Figure~\ref{fig:sample} shows the transfer performance from CHIFIR to PIFIR across budgets under two strong baselines. BatchBALD is highly unstable; a small change in budget can flip the model from almost perfect to completely broken. TAGCOS is more stable but still remains unreliable at small budgets.
Figure~\ref{fig:F1} shows the transfer performance from CHIFIR to PIFIR with our method. The left graph shows that starting from budget 5, F1 on PIFIR stays around 0.85–0.87 and additional labels provide marginal improvements, while CHIFIR performance remains stable. 
The right graph plots the domain gap \(\Delta\)F1 versus budget with 95\% confidence intervals. From budget = 4 onward, the gap is effectively closed (\(\Delta\)F1 \(\approx 0\) with overlapping confidence intervals), suggesting the selected subset suffices to eliminate the transfer gap.

As MIMIC-CXR is larger than CHIFIR, identifying informative target samples is less challenging for all models, even under low annotation budgets. The transfer performance from MIMIC-CXR to PIFIR across budgets is shown in Appendix \ref{sec:m->p}. 

\subsection{Ablation Study}

\begin{table}[h]
  \centering
  \small
  \renewcommand{\arraystretch}{1.05}
  \setlength{\tabcolsep}{5pt}
  \resizebox{\columnwidth}{!}{%
  \begin{tabular}{l|ccccc}
    \toprule
    \textbf{Method} & \textbf{Accuracy} & \textbf{F1-score} & \textbf{Precision} & \textbf{Recall} & \textbf{ROC-AUC} \\
    \midrule
    No RL          & 0.571 & 0.609 & 0.933 & 0.452 & 0.798 \\
    MI Only        & 0.262 & 0.000 & 0.000 & 0.000 & 0.475 \\
    Utility Only   & 0.786 & 0.866 & 0.806 & 0.935 & 0.833 \\
    \midrule
    \textbf{RADS} & \textbf{0.810} & \textbf{0.871} & \textbf{0.871} & \textbf{0.871} & \textbf{0.833} \\
    \bottomrule
  \end{tabular}
  }
  \caption{Ablation results under the hard transfer setting from CHIFIR to PIFIR with a labeling budget of 5.} 
  \label{tab:ablation}
\end{table}

To evaluate the effectiveness of each component in RADS, we conduct ablation studies as shown in Table \ref{tab:ablation}.
Replacing the RL sampler with a greedy selector (No RL) leads to a clear drop performance. Although this variant optimizes the same objective, it lacks the sequential decision-making needed to balance exploration and redundancy control. Selecting samples by BALD signal (MI Only) fails, implying that uncertainty-only criteria can favor noisy or out-of-distribution target examples under domain shift. Using the prior-aware utility (Utility Only) greatly improves results, confirming the benefit of class- and quality-aware selection, but remains below our method, highlighting the additional gains from discouraging redundant selections. Finally, RADS further improves over Utility Only, demonstrating that the RL sampler provides additional gains by learning a non-redundant, globally optimized subset selection policy rather than relying on pointwise ranking. 


\subsection{Selected Sample Quality Analysis}

We audit the quality of the selected target samples. Because RADS is trained and applied on the same unlabeled target pool, we clarify that the sampler never accesses target gold labels during optimization. 
Figure~\ref{fig:ratio} shows that in CHIFIR to PIFIR transfer, the source-trained pseudo labels are notably misaligned with the annotated labels, yet RADS still selects mostly true positives, which helps improve target performance. In MIMIC-CXR to PIFIR, the pseudo-label ratio better matches the annotated composition, consistent with smaller domain shift and improved calibration on the target domain.

\begin{figure}[h]
  \centering
  \includegraphics[width=0.95\linewidth]{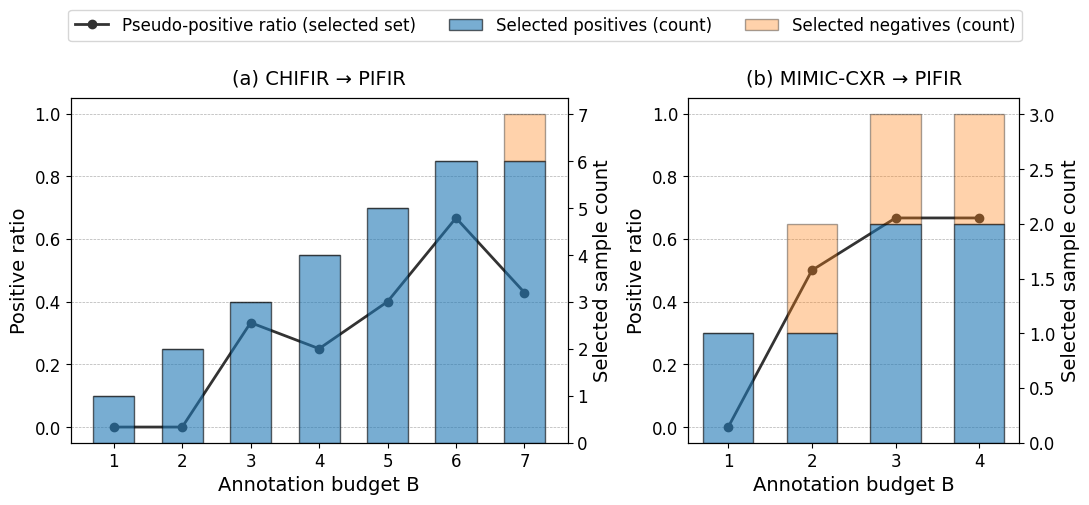}
  \caption{Selected sample analysis under CHIFIR to PIFIR (left) and MIMIC-CXR to PIFIR (right) transfer. The black line shows the pseudo-positive ratio predicted by the source-trained active learner, and the bars report the numbers of true positives and true negatives after manual annotation (blue/yellow).
  }
  \label{fig:ratio}
    \vspace{-0.5em}
\end{figure}

Figure~\ref{fig:PCA} visualizes the CHIFIR to PIFIR (left) and MIMIC-CXR to PIFIR (right) transfer, where our method selects 5 samples for adaptation. Before transfer learning, the decision boundary only captures the source dataset specific separation. After adding the selected target subset, the boundary rotates and shifts toward a direction that better reflects the class layout from both datasets.

\begin{figure}[t]
  \centering
\includegraphics[width=1\linewidth]{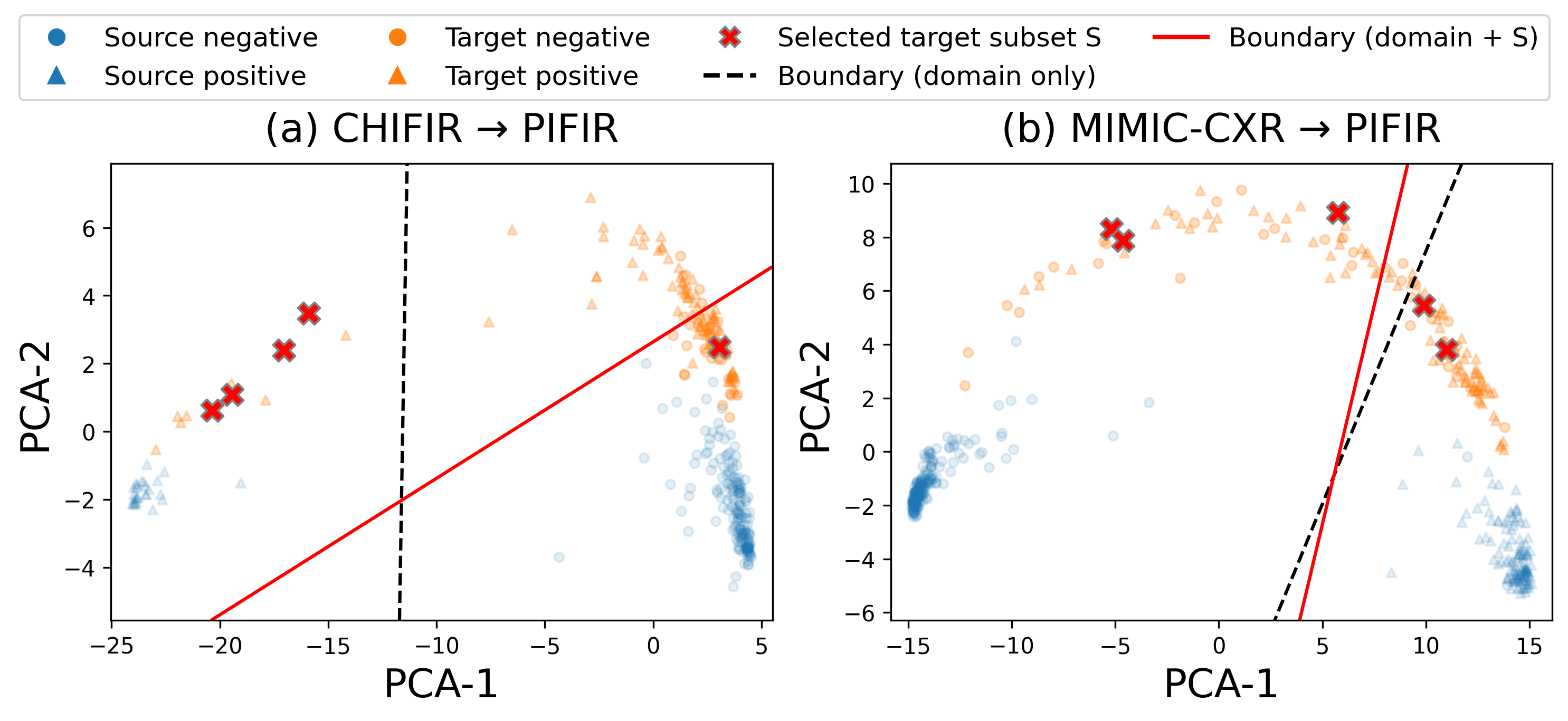}
  \caption{Transfer learning from CHIFIR to PIFIR (left) and MIMIC-CXR to PIFIR (right) with PCA projection of report embeddings.
  Red \(\times\) markers denote the selected PIFIR subset \(S\). The dashed black line shows the decision boundary learned from the source dataset only. The solid red line shows the boundary after augmenting with selected samples. Source = CHIFIR / MIMIC-CXR dataset, Target = PIFIR datatset.
  }
  \label{fig:PCA}
\end{figure}

\subsection{Efficiency and Budget-Aware Transfer}

\paragraph{Runtime Efficiency}
Our RL-based sampler in RADS is light-weight, requiring only a few seconds to produce a selection. Selecting 2 samples (MIMIC-CXR to PIFIR transfer) takes around 3 seconds, and selecting 5 samples (CHIFIR to PIFIR transfer) takes around 9 seconds. Given that annotating a single report takes about one minute, the selection overhead is negligible. Compared with full-shot transfer that labels the entire target training set (135 reports), RADS achieves comparable target performance with only 2 or 5 annotated reports.

\begin{figure}[h]
  \centering
  \includegraphics[width=0.49\linewidth]{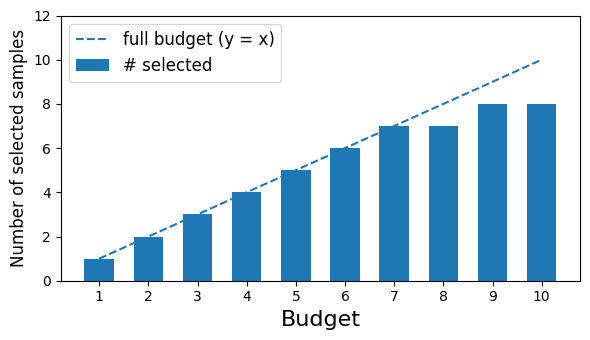}
  \includegraphics[width=0.49\linewidth]{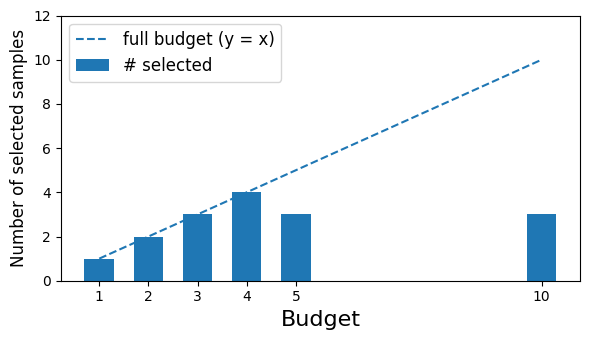}
  \caption{Number of selected PIFIR samples versus the annotation budget \(B\) when training on CHIFIR to PIFIR (left) and MIMIC-CXR to PIFIR (right).}
  \label{fig:num}
\end{figure}

\paragraph{Budget Utilization and Early Stopping}
As our RL-based sampler encodes budget progress (\(|S_t|/B\)) in the state, it can also provide guidance on how many samples are worth annotating during transfer. Figure~\ref{fig:num} shows the actual number of PIFIR samples selected as the budget increases. 
For CHIFIR to PIFIR transfer, the policy stops fully consuming the budget once \(B \ge 8\), consistent with our results showing good transfer performance already at \(B=5\).
For MIMIC-CXR to PIFIR transfer, the policy no longer uses the full budget when \(B \ge 5\), aligning with 
good target performance achieved with 
\(B=2\) labeled PIFIR samples.
This pattern is useful as in active learning it is important to know when it is time to stop adding samples. 

\subsection{Transfer Gap between Datasets} 

To explain why model transfer from MIMIC-CXR to PIFIR is easier than CHIFIR to PIFIR, we further analyze the distribution gap between these datasets.

We quantify overlap between datasets with a shared unigram--bigram vocabulary \cite{elangovan-etal-2021-memorization}. Coverage of PIFIR-test n-grams is higher for MIMIC-CXR to PIFIR than for CHIFIR to PIFIR (0.193 vs.\ 0.115).
The Jaccard similarity between source and target vocabularies is also higher for MIMIC-CXR to PIFIR than for CHIFIR to PIFIR (0.187 vs.\ 0.124). 
This suggests a smaller lexical domain shift between MIMIC-CXR and PIFIR, as it is illustrated by our empirical results fine-tuning MIMIC-CXR.
More detailed differences are analyzed in Appendix~\ref{sec:analysis}.


\section{Conclusion}

In this work, we studied transfer learning for disease detection under low-resource and class-imbalanced conditions. 
We proposed RADS, an RL-based sampler that jointly optimizes a prior-aware utility for class-mixture control and a diversity regularizer to avoid near-duplicate selections. Our approach improves model performance and adaptability across medical datasets compared to traditional sample selection strategies. We expect this approach to generalize to other transfer learning problems not only in clinical NLP but also in broader application domains, offering a promising direction for broader validation and impact. 


\section*{Limitations}

Despite demonstrating promising results, our approach has several limitations. First, the effectiveness of our RL-based sample selection heavily depends on the feedback provided by the active learner. This places high quality demands on the original gold dataset. 
Second, our formulation controls class mixture only through predicted priors and does not explicitly incorporate richer clinical knowledge. 
Third, our experiments focus on binary clinical disease detection with relatively small target pools and we have not yet validated RADS on larger-scale multi-class settings or on broader non-clinical transfer tasks. Fourth, more stable RL optimizers, improved uncertainty estimation, and better transfer-aligned validation strategies remain important directions for future work.


\section*{Acknowledgments}

This work was supported by the Australian Government through Medical Research Future Fund grant MRFCRI000188. The authors also thank the EINSTEIN Study Group for their support and insightful discussions.

\bibliography{custom}
\newpage
\appendix

\section{Algorithm Pseudocode}
\label{sec:algo}

We provide the pseudocode for the strategy of sample selection in our approach below.

\begin{algorithm}[h]
\label{alg:rl_select}
\small
\begin{algorithmic}[1]
\Require pool $\mathcal{U}_t$, budget $B$, episodes $N$, utility $u(\cdot)$, diversity weight $\lambda$
\Statex \hspace{\algorithmicindent} feature set $\mathcal{F}=\{\bar{\ell}(x),\mathrm{PE}(x),\widetilde{\mathrm{MI}}(x),|S|/B\}$, action set $\mathcal{A}=\{0,1\}$

\State $\textit{env}\gets\textsc{RLSampleSelectionEnv}(\mathcal{U}_t,\mathcal{F},u,B,\lambda)$
\State Initialize online network $Q_{\phi}$ and target network $\hat{Q}_{\phi}\gets Q_{\phi}$
\State Initialize replay buffer $\mathcal{D}\gets \emptyset$
\State Initialize exploration rate $\epsilon$

\For{$\textit{episode}=1$ \textbf{to} $N$}
  \State $s\gets\textit{env.reset()}$; \quad $\textit{done}\gets \textbf{false}$
  \While{\textbf{not} $\textit{done}$}
    \State $a\gets \Call{EpsGreedy}{Q_{\phi},s,\epsilon}$
    \State $(s',r,\textit{done})\gets \textit{env.step}(a)$
    \State Add $(s,a,r,s',\textit{done})$ to $\mathcal{D}$
    \If{$|\mathcal{D}|\ge M$} \Comment{$M$ is minibatch size}
      \State $\mathcal{M}\gets \Call{SampleMinibatch}{\mathcal{D},M}$
      \State \Call{UpdateNets}{$Q_{\phi},\hat{Q}_{\phi},\mathcal{M}$}
    \EndIf
    \State $s\gets s'$
  \EndWhile
  \If{$\textit{episode}\bmod K_{\text{upd}} = 0$}
    \State $\hat{Q}_{\phi}\gets Q_{\phi}$
  \EndIf
  \State $\epsilon \gets \Call{DecayEps}{\epsilon}$
\EndFor

\Statex \textbf{Selection (Greedy Policy)}
\State $\mathcal{S}\gets\emptyset$; \quad $s\gets\textit{env.reset()}$; \quad $\textit{done}\gets \textbf{false}$
\While{\textbf{not} $\textit{done}$}
  \State $id\gets \textit{env.currentId()}$
  \State $a\gets \arg\max_{a'\in\mathcal{A}} Q_{\phi}(s,a')$
  \State $(s',\_,\textit{done})\gets \textit{env.step}(a)$
  \If{$a=1$}
    \State $\mathcal{S}\gets \mathcal{S}\cup\{id\}$
  \EndIf
  \State $s\gets s'$
\EndWhile
\State \Return $\mathcal{S}$
\end{algorithmic}
\end{algorithm}

\begin{table*}[t]
  \centering
  \small
  \renewcommand{\arraystretch}{1}       
  \setlength{\tabcolsep}{4.5pt}
  \setlength{\heavyrulewidth}{1.2pt}  
  \resizebox{0.7\textwidth}{!}{
  \begin{tabular}{ll|cccc|cccc}
    \toprule
    \multicolumn{2}{c|}{\textbf{Transfer Learning to CHIFIR}} 
      & \multicolumn{4}{c|}{\textbf{Performance on CHIFIR}} 
      & \multicolumn{4}{c}{\textbf{Performance on PIFIR}} \\
    \multicolumn{1}{c}{\textbf{Strategy}} 
      & \multicolumn{1}{c|}{\textbf{Datasets}}
      & Acc & F1 & P & R 
      & Acc & F1 & P & R \\
    \midrule
    Baseline
      & CHIFIR
      & 0.942 & 0.824 & 0.778 & 0.875
      & --   & --   & --   & -- \\ 
    \midrule
    \multirow{1}{*}{Zero-shot} 
      & PIFIR
      & 0.154 & 0.267 & 0.154 & 1.000
      & 0.714 & 0.812 & 0.788 & 0.839\\ 

    \midrule
    \multirow{1}{*}{Full-shot}
      & PIFIR + CHIFIR
      & 0.904 & 0.615 & 0.800 & 0.500
      & 0.857 & 0.900 & 0.931 & 0.871\\

    \bottomrule
  \end{tabular}
  }
  \caption{
  Performance comparison of transfer learning from PIFIR to CHIFIR under zero-shot transfer and full-shot transfer.
  }
  \label{tab:baseline-2}
\end{table*}

\begin{table*}[h]
  \centering
  \scriptsize
  \renewcommand{\arraystretch}{1.15}
  \setlength{\tabcolsep}{4pt}
  \resizebox{0.95\textwidth}{!}{%
    \begin{tabular}{c|ccccc|ccccc|cc}
      \toprule
      \multicolumn{13}{c}{\textbf{Knowledge Transfer from PIFIR to CHIFIR}} \\
      \midrule
      \multirow{2}{*}{\textbf{Strategy}}
        & \multicolumn{5}{c|}{\textbf{Performance on CHIFIR}}
        & \multicolumn{5}{c|}{\textbf{Performance on PIFIR}}
        & \multicolumn{2}{c}{\textbf{Transfer Gap}} \\
      & Accuracy & F1-score & Precision & Recall & ROC-AUC
        & Accuracy & F1-score & Precision & Recall & ROC-AUC
        & $\Delta$F1 & 95\% CI \\
      \midrule

      Random
        & 0.838 & 0.167 & 0.250 & 0.125 & 0.716
        & 0.752 & 0.791 & 0.916 & 0.729 & 0.837
        & -- & -- \\
      Uncertainty
        & 0.788 & 0.267 & 0.286 & 0.250 & 0.656
        & 0.905 & 0.938 & 0.909 & 0.968 & 0.880
        & 0.671 & [0.387, 0.967] \\
      Diversity
        & 0.154 & 0.267 & 0.154 & 1.000 & 0.591
        & 0.738 & 0.849 & 0.738 & 1.000 & 0.883
        & 0.584 & [0.409, 0.752] \\
      LM-DPP
        & 0.154 & 0.267 & 0.154 & 1.000 & 0.489
        & 0.738 & 0.849 & 0.738 & 1.000 & 0.815
        & 0.583 & [0.409, 0.752] \\
      TAGCOS
        & 0.769 & 0.143 & 0.167 & 0.125 & 0.545
        & 0.929 & 0.954 & 0.912 & 1.000 & 0.827
        & 0.811 & [0.529, 0.986] \\
      BatchBALD
        & 0.846 & 0.000 & 0.000 & 0.000 & 0.486
        & 0.714 & 0.793 & 0.852 & 0.742 & 0.742
        & 0.793 & [0.654, 0.897] \\
      \rowcolor{gray!16}
      \textbf{RADS}
        & \textbf{0.865} & \textbf{0.632} & \textbf{0.545} & \textbf{0.750} & \textbf{0.858}
        & \textbf{0.881} & \textbf{0.921} & \textbf{0.906} & \textbf{0.935} & \textbf{0.865}
        & \textbf{0.289} & \textbf{[0.075, 0.599]} \\
      \bottomrule
    \end{tabular}
  }
  \caption{Transfer learning performance from PIFIR to CHIFIR with 8 samples selected in CHIFIR under different sample selection strategies. $\Delta$F1 = F1(PIFIR) $-$ F1(CHIFIR). CI = Confidence Interval.}
  \label{tab:pc_transfer}
\end{table*}

\section{Reproducibility}
\label{sec:repro}

All experiments are conducted on a single NVIDIA A100 GPU. 
Key fine-tuning settings are: epochs = 15, learning rate = $2\times10^{-5}$, batch size = 8, max sequence length = 512, weight decay = 0.01, and early stopping with a patience of 3 epochs.

For uncertainty estimation, we use MC dropout with the number of stochastic forward passes set to $K=10$.
In RADS, 
we train a dueling DQN sampler for 300 episodes with $\epsilon$-greedy exploration, decaying $\epsilon$ from 1.0 to 0.05 with a multiplicative factor of 0.995. We use an experience replay buffer of size 10000 and start network updates once at least one minibatch is available (batch size = 64). Both the online and target networks are optimized with Adam (learning rate = $10^{-4}$) and discount factor $\gamma=0.95$, and the target network is synchronized every 10 episodes. $\rho=0.9$. The reward is defined as the prior-aware utility minus a diversity penalty computed from the nearest-neighbor distance in the mean log-probability space $\bar{\ell}(x)$, with $\lambda=0.01$. 

\paragraph{Dataset Split}
Each dataset is split into training, validation, and test sets (around 70\%, 10\%, and 20\%), preserving the original class balance. Table~\ref{tab:data_stats} shows the number of positive and negative samples in each split. 

\begin{table}[h]
  \centering
  \small
  \renewcommand{\arraystretch}{1}
  \resizebox{\linewidth}{!}{
  \begin{tabular}{l|ccc|ccc|ccc}
    \toprule
    \multirow{2}{*}{\textbf{Split}} 
      & \multicolumn{3}{c|}{\textbf{CHIFIR}} 
      & \multicolumn{3}{c|}{\textbf{MIMIC-CXR}}
      & \multicolumn{3}{c}{\textbf{PIFIR}} \\
      & Total & P & N & Total & P & N & Total & P & N \\
    \midrule
    Train & 196 & 27 & 169 & 320 & 124 & 196 & 135 & 92 & 43 \\
    Dev   & 35  & 5  & 30  & 74  & 29  & 45  & 24  & 16 & 8  \\
    Test  & 52  & 8  & 44  & 99  & 38  & 61  & 42  & 31 & 11 \\
    \bottomrule
  \end{tabular}}
  \caption{
    Class distribution for CHIFIR, MIMIC-CXR, and PIFIR across train, development, and test sets. P = the number of positive reports, N = the number of negative reports.}
  \label{tab:data_stats}
\end{table}

\section{Prompt-guided LLM Selection Baseline}
\label{app:llm_selection}

\begin{table}[h]
\centering
\small
\begin{tabular}{c c c c c}
\toprule
Num & Accuracy & F1-score & Precision & Recall \\
\midrule
1 & 0.261 & 0.000 & 0.000 & 0.000 \\
2 & 0.285 & 0.062 & 1.000 & 0.032 \\
4 & 0.261 & 0.000 & 0.000 & 0.000 \\
8 & 0.642 & 0.681 & 1.000 & 0.516 \\
16 & 0.571 & 0.591 & 1.000 & 0.419 \\
32 & 0.942 & 0.800 & 0.857 & 0.750 \\
\bottomrule
\end{tabular}
\caption{Prompt-guided LLM selection results for CHIFIR to PIFIR transfer. Num is the number of selected PIFIR reports.}
\label{tab:llm_select_chifir_pifir}
\end{table}

We compare RADS against a prompt-guided LLM selection baseline inspired by recent LLM-based data selection methods \cite{jeong2025llmselect}. In this baseline, we use an open-source medical LLM (OpenBioLLM-8B\footnote{\url{https://huggingface.co/aaditya/Llama3-OpenBioLLM-8B}}) to score each unlabeled target report according to its estimated usefulness for training the IFI classifier, and then select the top-k reports for annotation.

Table \ref{tab:llm_select_chifir_pifir} reports results for transfer from CHIFIR to PIFIR under different annotation budgets. We observe that the LLM-guided baseline is highly unstable in the ultra-low-budget regime: at budgets 1 and 4, it completely fails to recover positive target performance, and at budget 2 it yields only a marginal F1 improvement. Performance improves at larger budgets, suggesting that the LLM can identify some useful reports when more annotations are allowed. However, the overall behavior remains much less reliable than RADS in the low-budget regime that is central to our study.

\section{Transfer Performance from PIFIR to CHIFIR}
\label{sec:p->c}

Table~\ref{tab:baseline-2} summarizes the baseline transfer performance. The zero-shot model trained on PIFIR exhibits severe performance degradation on CHIFIR, achieving very low accuracy and F1, indicating a substantial domain shift. In particular, the model tends to over-predict the positive class on CHIFIR (high recall but very low precision), which suggests that the decision boundary learned from PIFIR does not directly generalize to CHIFIR. When CHIFIR data are available for adaptation (full-shot), incorporating CHIFIR supervision mitigates this shift and improves target-domain performance, demonstrating that a small amount of target data is critical for reliable transfer.

We next evaluate whether our data selection strategy can maximize the benefit of limited target supervision. Table~\ref{tab:pc_transfer} reports transfer learning results when only 8 CHIFIR samples are selected for fine-tuning under different sampling strategies. Across all baselines, we observe that other strategies are insufficient to bridge the transfer gap: they either fail to improve CHIFIR F1 or produce unstable behavior.
In contrast, RADS achieves the strongest transfer performance on CHIFIR, yielding the best overall target metrics (Accuracy/F1/ROC-AUC) while maintaining high performance on the source domain. Importantly, RADS also produces the smallest transfer gap ($\Delta$F1) among compared methods, indicating that the selected CHIFIR samples lead to more effective adaptation without sacrificing the knowledge learned from PIFIR. The confidence interval of $\Delta$F1 further suggests that RADS provides a more reliable and stable reduction of the transfer discrepancy compared to alternative selection strategies.

\begin{figure*}[h]
  \includegraphics[width=\textwidth]{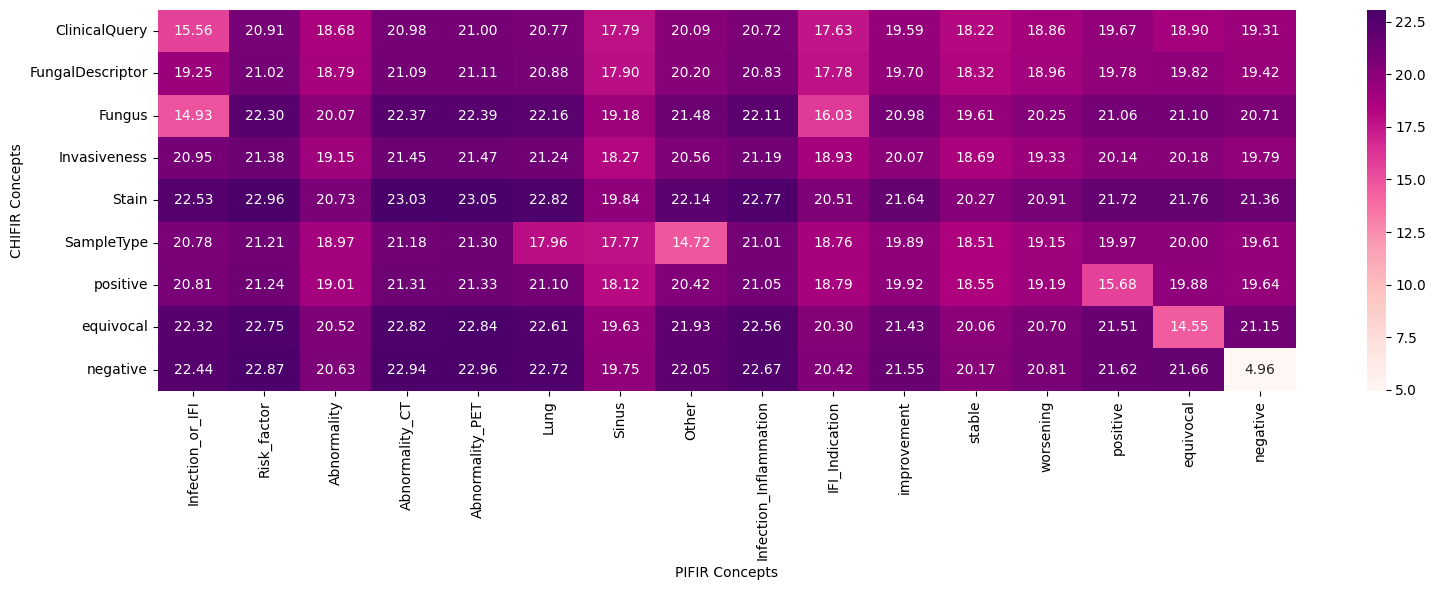}
  \caption{Concept-level KL divergence from CHIFIR to PIFIR.}
  \label{fig:C-P}
\end{figure*}

\section{Robustness under Imbalanced Sampling}
\label{sec:robust}
We evaluate the robustness of our sampling strategy under imbalanced sampling. For transfer learning from CHIFIR to PIFIR, We randomly select 5 samples from PIFIR with different positive to negative ratios. Each setting is repeated five times to obtain stable results. Figure~\ref{fig:pn_ratio} shows the results. The best performance occurs when the positive to negative ratio is 1.00:0.00, which matches the class ratio selected by our method. This happens because CHIFIR has many more negative cases. Prioritizing positive target samples helps counter this imbalance and narrow the target-domain class distribution gap during transfer.

\begin{figure}[h]
  \centering
  \includegraphics[width=0.8\linewidth]{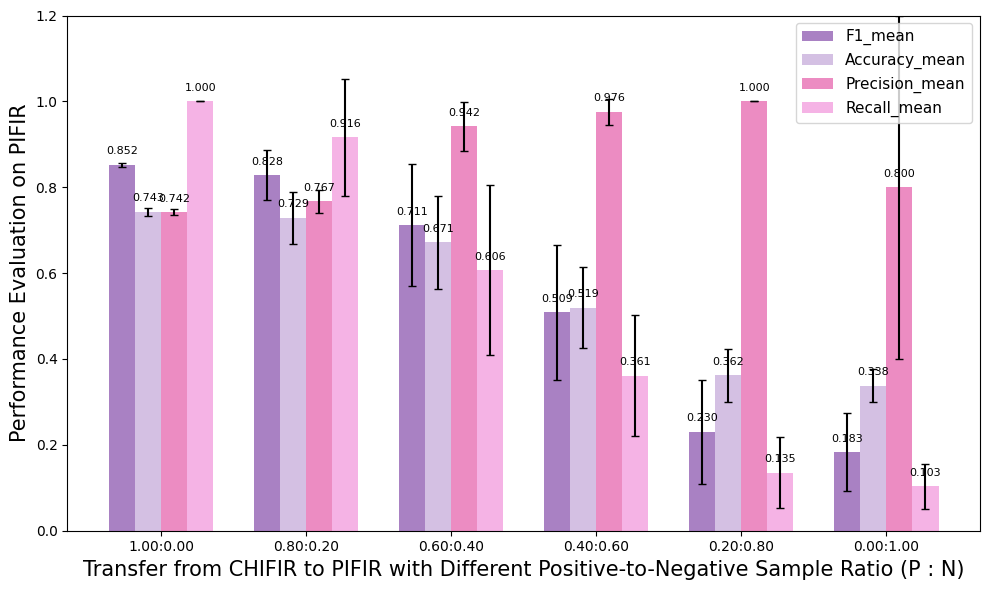} \hfill
  \caption{Class imbalance analysis of positive to negative sample ratios for CHIFIR to PIFIR transfer. Bars show mean values and black lines indicate variance.}

  \label{fig:pn_ratio}
\end{figure}

\begin{table*}[t]
\centering
\small
\setlength{\tabcolsep}{5.0pt}
\renewcommand{\arraystretch}{1}
\begin{tabular}{l r|l r|l r}
\toprule
\multicolumn{2}{c|}{\textbf{CHIFIR}} & \multicolumn{2}{c|}{\textbf{PIFIR}} & \multicolumn{2}{c}{\textbf{MIMIC-CXR}} \\
\textbf{Word} & \textbf{TF-IDF} & \textbf{Word} & \textbf{TF-IDF} & \textbf{Word} & \textbf{TF-IDF} \\
\midrule
cells       & 18.405196 & uptake    & 17.527355 & chest      & 33.730506 \\
fluid       & 12.311106 & ct        & 16.972962 & pneumonia  & 30.779783 \\
bronchial   & 11.999560 & fdg       & 14.968679 & right      & 27.327142 \\
description & 11.998803 & pet       & 11.530773 & left       & 24.813217 \\
biopsy      & 11.490593 & right     &  9.951929 & pleural    & 24.584867 \\
tissue      & 10.625458 & marrow    &  8.764460 & effusion   & 22.224288 \\
specimen    &  9.652244 & activity  &  8.720572 & pulmonary  & 21.700909 \\
lung        &  9.026040 & disease   &  8.699531 & lung       & 21.272838 \\
washings    &  9.022390 & left      &  8.493931 & comparison & 21.251147 \\
cell        &  8.798129 & findings  &  8.308830 & findings   & 20.950609 \\
\bottomrule
\end{tabular}
\caption{Top 10 terms with the highest TF--IDF scores in CHIFIR, PIFIR and MIMIC-CXR.}
\label{tab:tfidf}
\end{table*}

\begin{figure*}[h]
  \includegraphics[width=\textwidth]{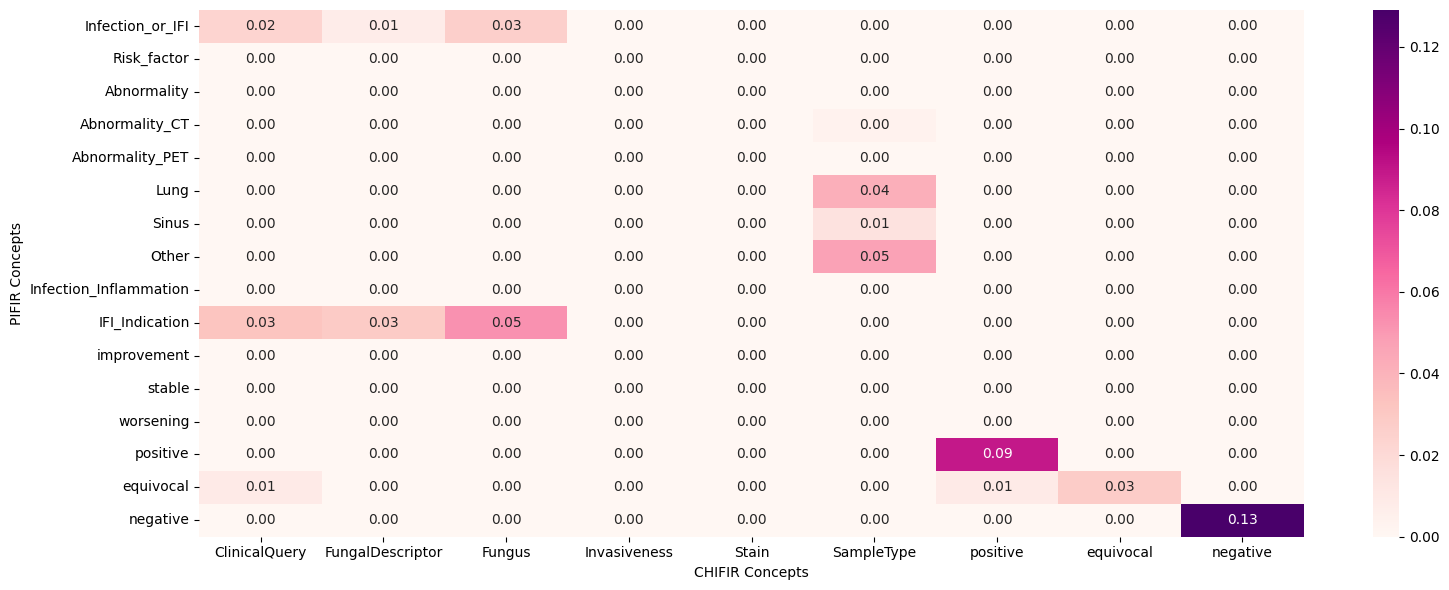}
  \caption{Jaccard similarity heatmap between CHIFIR and PIFIR concepts.}
  \label{fig:appendix}
\end{figure*}

\section{Learning Curves for MIMIC-CXR to CHIFIR Transfer under Varying Budgets}
\label{sec:m->p}
Figure~\ref{fig:sample-m} shows the performance from MIMIC-CXR to PIFIR across budgets under two baselines and Figure~\ref{fig:F1-m} (left) shows our methods' performance. MIMIC-CXR is larger and the zero-shot baseline is already good. Therefore, the headroom for improvement is limited, and our method is similar to other baselines.
Figure~\ref{fig:F1-m} (right) plots the domain gap \(\Delta\)F1 against budget with 95\% confidence intervals. Across budgets, the point estimates stay close to zero and the confidence intervals largely overlap, indicating a small residual gap and no clear separation between budgets in this ultra-low-resource setting.

\begin{figure}[h]
  \centering
  \includegraphics[width=0.49\linewidth]{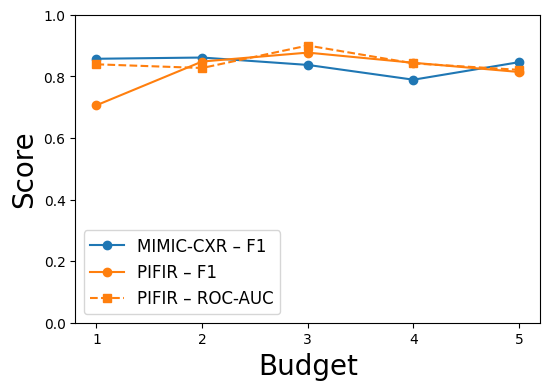} 
  \includegraphics[width=0.49\linewidth]{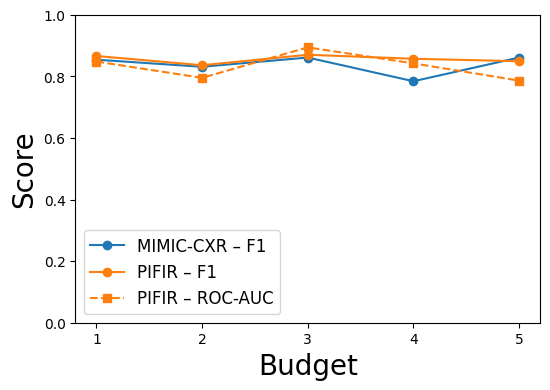}
  \caption{Transfer from MIMIC-CXR to PIFIR 
  under baselines BatchBALD (left) and TAGCOS (right).}
  \label{fig:sample-m}
\end{figure}

\begin{figure}[h]
  \centering
  \includegraphics[width=0.47\linewidth]{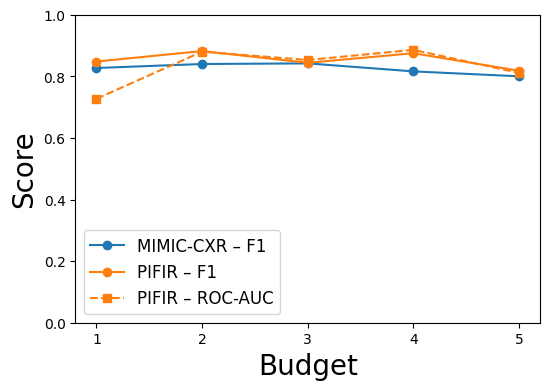}
  \includegraphics[width=0.49\linewidth]{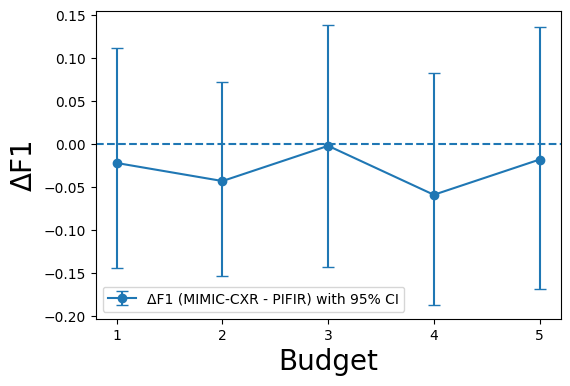}
  \caption{Transfer from MIMIC-CXR to PIFIR 
  under our method RADS.}
  \label{fig:F1-m}
\end{figure}
\section{Differences Analysis in CHIFIR, PIFIR, and MIMIC-CXR Datasets}
\label{sec:analysis}

CHIFIR contains 283 reports from 201 patients, with an average report length of 1,353 characters. 
PIFIR contains 201 reports from 156 patients, with an average report length of 1,809 characters. 
The MIMIC-CXR subset contains 493 reports from 290 patients, with a shorter average report length of 677 characters.


We first compute TF–IDF scores within each corpus and compare the top 10 highest-scoring terms between CHIFIR, PIFIR and MIMIC-CXR as shown in Table \ref{tab:tfidf}.
CHIFIR is dominated by pathology and specimen-centric language (e.g., cells, fluid, bronchial, biopsy, tissue, specimen), reflecting cytology/histopathology reporting that emphasizes sample type and microscopic description rather than imaging observations. 
PIFIR is characterized by PET-CT and metabolic-imaging terminology (e.g., uptake, FDG, PET, CT, activity), as well as systemic disease descriptors (e.g., marrow, disease), consistent with PET-driven assessment of metabolic activity and whole-body involvement.
MIMIC-CXR is dominated by chest radiography vocabulary and common pulmonary findings (e.g., chest, pneumonia, pleural, effusion, pulmonary, lung), reflecting the focus of X-ray reports on thoracic anatomy and acute cardiopulmonary abnormalities.
Overall, these TF--IDF profiles highlight substantial modality- and workflow-driven lexical shifts between these datasets, motivating domain-adaptive transfer methods that can operate under pronounced vocabulary mismatch. Figure \ref{fig:wordclouds} also shows the transfer gap.

For the CHIFIR and PIFIR datasets, expert annotators also provided span-level annotation of concepts relevant to disease detection. Concept annotations in CHIFIR and PIFIR Datasets are listed in Table~\ref{tab:CHIFIR_stat} and Table~\ref{tab:PIFIR_stat}. The CHIFIR dataset reports 1,155 concepts, and the PIFIR dataset has 3,194 concepts. The two corpora serve different clinical niches. CHIFIR comes from cytology and histopathology notes and therefore focuses on microbiology terms such as \texttt{FungalDescriptor} and \texttt{Stain}. PIFIR is built from PET–CT reports and centres on imaging findings and risk factors, for example \texttt{Abnormality\_CT} and \texttt{Risk\_factor}.

\begin{table*}[h]
\centering
\small
\setlength{\tabcolsep}{6pt}
\renewcommand{\arraystretch}{1.15}
\begin{tabular}{p{0.12\textwidth} p{0.83\textwidth}}
\toprule
\textbf{Dataset} & \textbf{Example report excerpt (de-identified)} \\
\midrule
\textbf{CHIFIR} &
\textit{""R groin LN biopsy"". Please note specimen has been received fresh and fragments have been sent to flow cytometry at 1225 on XXXXXX by XX/XXX. Two tan wispy cores 4 and 5 mm in length. A1. (dl:oze) 
""R groin LN biopsy"". A tan core, 4mm in length with multiple fragments up to 1mm. A1. (dl/kr) } \\
\addlinespace
\textbf{PIFIR} &
\textit{PET/CT technique: Scanning was performed encompassing the base of skull to upper thighs on a PET/CT scanner (GE 690 with time-of-flight). A contemporaneous low dose non-contrast multislice CT scan was performed for anatomic correlation and attenuation correction. Uptake time=70 minutes. BSL=<7mmol/L.  } \\
\addlinespace
\textbf{MIMIC-CXR} &
\textit{AP upright and lateral views the chest were provided. Mild left basal
 atelectasis. Lungs are otherwise clear.  No signs of pneumonia or edema. No large effusion or pneumothorax. Cardiomediastinal silhouette is normal. Bony structures are intact. No free air below the right hemidiaphragm.} \\
\bottomrule
\end{tabular}
\caption{Representative (de-identified) report excerpts from each dataset.}
\label{tab:report-examples}
\end{table*}

To quantify overlap, we compute the Jaccard Similarity between the concept vocabularies:

\begin{equation}
  \label{eq:js}
  J(A,B)=\frac{|A\cap B|}{|A\cup B|}
\end{equation}

where \(A\) and \(B\) are the sets of surface forms in CHIFIR and PIFIR, respectively. Figure \ref{fig:appendix} plots the resulting heatmap. Although both datasets include the classification terms \texttt{positive}, \texttt{equivocal} and \texttt{negative}, their lexical realizations share little common ground, so the Jaccard scores remain low.

\begin{table}[h]
  \centering
  \small
  \resizebox{0.99\linewidth}{!}{%
    \begin{tabular}{l c c c}
      \hline
      \textbf{Concept} & \textbf{Count} & \textbf{Unique} & \textbf{Diversity} \\
      \hline
      \texttt{Infection\_or\_IFI} & 279 &174 & 0.62 \\
      \texttt{Risk\_factor} & 429 &179 & 0.42 \\
      \texttt{Abnormality} &  46 & 24 & 0.52 \\
      \texttt{Abnormality\_CT} & 460 &204 & 0.44 \\
      \texttt{Abnormality\_PET} & 470 &224 & 0.48 \\
      \texttt{Lung} & 372 & 36 & 0.10 \\
      \texttt{Sinus} &  19 &  4 & 0.21 \\
      \texttt{Other} & 189 & 92 & 0.49 \\
      \texttt{Infection\_Inflammation} & 354 &103 & 0.29 \\
      \texttt{IFI\_Indication} &  37 & 21 & 0.57 \\
      \texttt{improvement} & 115 & 51 & 0.44 \\
      \texttt{stable} &  29 & 16 & 0.55 \\
      \texttt{worsening} &  55 & 34 & 0.62 \\
      \texttt{positive} & 124 & 33 & 0.27 \\
      \texttt{equivocal} & 129 & 68 & 0.53 \\
      \texttt{negative} &  87 & 23 & 0.26 \\
      \hline
    \end{tabular}%
  }
  \caption{Summary statistics for the IFI-related concepts in the PIFIR dataset.}
  \label{tab:PIFIR_stat}
\end{table}

\begin{table}[h]
  \centering
  \small
  \resizebox{0.85\linewidth}{!}{%
    \begin{tabular}{l c c c}
      \hline
      \textbf{Concept} & \textbf{Count} & \textbf{Unique} & \textbf{Diversity} \\
      \hline
      \texttt{ClinicalQuery}    &  68 & 43 & 0.63 \\
      \texttt{FungalDescriptor} & 294 & 86 & 0.29 \\
      \texttt{Fungus}           & 106 & 19 & 0.18 \\
      \texttt{Invasiveness}     &  39 & 27 & 0.69 \\
      \texttt{Stain}            & 172 & 16 & 0.09 \\
      \texttt{SampleType}       & 198 & 64 & 0.32 \\
      \texttt{positive}         & 118 & 40 & 0.34 \\
      \texttt{equivocal}        &   8 &  6 & 0.75 \\
      \texttt{negative}         & 152 & 12 & 0.08 \\
      \hline
    \end{tabular}%
  }
  \caption{Summary statistics for the IFI-related concepts in the CHIFIR dataset.}
  \label{tab:CHIFIR_stat}
\end{table}

To quantify directional lexical divergence, we compute the KL divergence on the concept level. For each concept, \(P\) and \(Q\) denote the distributions of surface forms in PIFIR and CHIFIR, respectively.  
Both are smoothed over the combined vocabulary \(\mathcal{V} = \mathcal{V}_{\text{PIFIR}} \cup \mathcal{V}_{\text{CHIFIR}}\) with a small \(\varepsilon\) to avoid zeros.

\begin{equation}
  \label{eq:kl}
  KL(P \,\|\, Q)
  = \sum_{v \in \mathcal{V}}
    P(v)\,\log\frac{P(v)}{Q(v)},
\end{equation}

\noindent
where
\begin{equation}
  \label{eq:kl_p}
  P(v) =
  \frac{\mathrm{count}_{\text{PIFIR}}(v) + \varepsilon}
       {\sum_{u \in \mathcal{V}} \mathrm{count}_{\text{PIFIR}}(u)
        + \varepsilon|\mathcal{V}|}.
\end{equation}

We compute $\mathrm{KL}(\text{CHIFIR} \| \text{PIFIR})$ and visualize the results as a heatmap (Figure \ref{fig:C-P}).  
In KL divergence, larger values indicate that greater mismatch between the two datasets.
Even for shared classification terms (\texttt{positive}, \texttt{equivocal}, \texttt{negative}), the divergence values remain large.  
This suggests that the two datasets differ systematically in how their concepts are expressed, not simply in whether specific words occur.

\paragraph{Representative Report Examples}

To qualitatively illustrate domain- and modality-specific language, we provide representative (de-identified) report excerpts from each corpus in Table~\ref{tab:report-examples}.

\end{document}